\documentclass{article}

\usepackage{PRIMEarxiv}

\usepackage[utf8]{inputenc} 
\usepackage[T1]{fontenc}    
\usepackage{hyperref}       
\usepackage{url}            
\usepackage{booktabs}       
\usepackage{amsfonts}       
\usepackage{nicefrac}       
\usepackage{microtype}      
\usepackage{lipsum}
\usepackage{fancyhdr}       
\usepackage{graphicx}       
\graphicspath{{media/}}     
\usepackage{fontawesome}
\usepackage{subfigure}
\usepackage{multirow}
\usepackage{multicol}
\usepackage{xcolor}
\usepackage{amsmath}
\usepackage{pgfplots}
\usepackage{comment}
\usepackage{subcaption}
\usepackage{graphicx}
\usepackage{float}

\pgfplotsset{compat=1.18}

\title{CCI3.0-HQ: a large-scale Chinese dataset of high quality designed for pre-training large language models}

\author{Liangdong Wang\thanks{Core contributors with equal contributions.} , Bo-Wen Zhang$^*$, Chengwei Wu$^*$, Hanyu Zhao$^*$, \\ \textbf{Xiaofeng Shi, Shuhao Gu, Jijie Li, Quanyue Ma,  TengFei Pan, Guang Liu\thanks{Project Lead, the corresponding author, contact \url{liuguang@baai.ac.cn}}} \\ \\
Beijing Academy of Artificial Intelligence(BAAI)
}

\begin{document}
\maketitle

\begin{abstract}
We present CCI3.0-HQ\footnote{\url{https://huggingface.co/datasets/BAAI/CCI3-HQ}}, a high-quality 500GB subset of the Chinese Corpora Internet 3.0 (CCI3.0)\footnote{\url{https://huggingface.co/datasets/BAAI/CCI3-Data}}, developed using a novel two-stage hybrid filtering pipeline that significantly enhances data quality. To evaluate its effectiveness, we trained a 0.5B parameter model from scratch on 100B tokens across various datasets, achieving superior performance on 10 benchmarks in a zero-shot setting compared to CCI3.0, SkyPile, and WanjuanV1. The high-quality filtering process effectively distills the capabilities of the Qwen2-72B-instruct model into a compact 0.5B model, attaining optimal F1 scores for Chinese web data classification. We believe this open-access dataset will facilitate broader access to high-quality language models.
\end{abstract}


\section{Introduction}

The success of Large Language Models (LLMs) \cite{dubey2024llama3herdmodels}\cite{yang2024qwen2technicalreport} is primarily attributed to the availability of extensive, high-quality pre-training corpora, which underpin their foundational knowledge and reasoning capabilities for a variety of tasks, from creative writing to complex problem-solving. Among them, the Open-source datasets, such as The Pile\cite{gao2020pile800gbdatasetdiverse} and Common Crawl\cite{commoncrawl}, have been instrumental in propelling LLM development, fostering collaboration and establishing benchmarks for innovation. 

Existing Researchers focus more on scaling high-quality data. Recently the demand for pre-training data has exceeded 10 trillion tokens \cite{dubey2024llama3herdmodels}\cite{qwen2.5}\cite{nvidia2024nemotron4340btechnicalreport}, underscoring two key trajectories in English pre-training: scaling data and improving its quality. Open-source datasets have rapidly expanded, evolving from collections like the Pile(825GB) to larger datasets such as FineWeb(15TB)\cite{penedo2024finewebdatasetsdecantingweb}, which draw extensively from Common Crawl. Simultaneously, the focus has shifted from rule-based filtering methods, as seen in early projects like Redpajama\cite{together2023redpajama}, to model-driven approaches exemplified by FineWeb-Edu\cite{penedo2024finewebdatasetsdecantingweb}. 

Despite the rapid advancement of English open-source datasets, Chinese data remains significantly underrepresented on the global web. Existing open-source Chinese datasets, such as WuDao \cite{YUAN202165}, SkyPile150B \cite{wei2023skywork}, and WanjuanV1 \cite{he2023wanjuancomprehensivemultimodaldataset}, are constrained in scale due to a scarcity of Chinese data sources online. Furthermore, there is limited research focused on improving quality classification for Chinese web data, resulting in suboptimal data quality. These challenges present substantial barriers to the development of high-performance Chinese language models, underscoring the urgent need for more effective data filtering and quality classification methodologies.

To address the identified challenges, we present CCI3.0-HQ, a large-scale Chinese pre-training dataset created through a two-stage hybrid filtering strategy: Fundamental Processing and High-Quality Processing. The fundamental stage encompasses standard web data curation practices, including safety filtering, text extraction, deduplication, and initial quality assessment using a basic model score. The second stage further enhances data quality by employing Qwen2-72B-Instruct \cite{yang2024qwen2technicalreport} to identify high-quality samples, resulting in a training set of 140k samples and a testing set of 14k samples. Our analysis shows that these annotations consistently align with GPT-4 annotations at approximately 80\%. Consequently, we train a 0.5B quality classifier on the 140k training samples to efficiently filter CCI3.0, producing a high-quality dataset.

To evaluate the impact of our dataset on training LLMs, we conducted a series of experiments using a 0.5B model trained from scratch on a 100B token with a specific data mix, assessing its performance on 10 benchmarks under a zero-shot setting. The extensive results demonstrate that CCI3.0-HQ significantly outperforms competing Chinese datasets, such as SkyPile and WanjuanV1. Additionally, our proposed quality classifier \textit{classifier}$_{\textit{CCI3.0-HQ}}$\footnote{\url{https://huggingface.co/BAAI/CCI3-HQ-Classifier}} achieved superior performance, surpassing the \textit{classifier}$_{\textit{FineWeb-edu}}$\footnote{\url{https://huggingface.co/HuggingFaceFW/fineweb-edu-classifier}}, \textit{classifier}$_{\textit{IndustryCorpus2}}$\footnote{\url{https://huggingface.co/BAAI/IndustryCorpus2_DataRater}}, and \textit{classifier}$_{\textit{ChineseWebText}}$\footnote{\url{https://huggingface.co/CASIA-LM/ChineseWebText-fasttext}} in terms of F1 score.

In summary, our major contributions are as follows:

\begin{itemize} 
\item We present CCI3.0-HQ, a groundbreaking 500GB Chinese pre-training dataset that leverages a sophisticated hybrid quality filtering methodology to enhance data integrity. 
\item We conduct rigorous experimental evaluations, demonstrating that CCI3.0-HQ substantially outperforms the original CCI3.0 dataset and other prominent open-source Chinese corpora, thereby establishing new benchmarks for performance. 
\item We introduce and open-source the CCI3-HQ classifier, an advanced quality classification tool that significantly improves data selection processes in LLM training. 
\end{itemize}

\section{Pipeline}

\begin{figure}
    \centering
    \begin{minipage}[t]{0.9\linewidth}
        \centering
        \includegraphics[width=\linewidth]{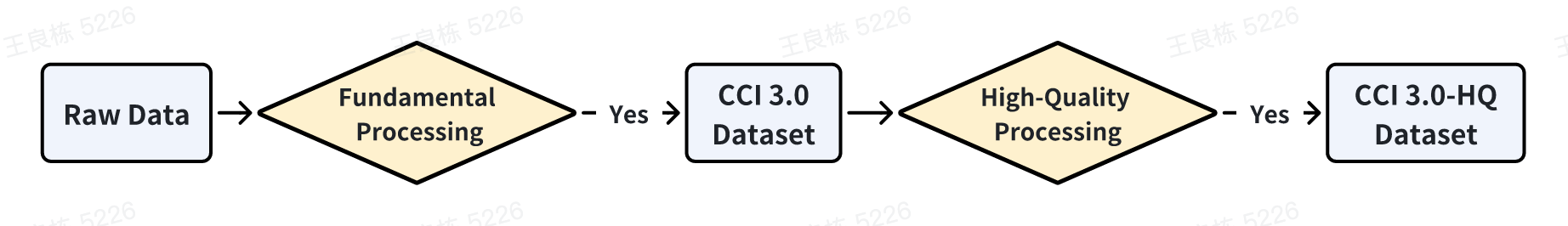}
    \end{minipage}
    \caption{Dataset Curation Pipeline}
    \label{fig:pipeline}
\end{figure}

As illustrated in Figure \ref{fig:pipeline}, the data processing pipeline comprises two primary phases: \textbf{Fundamental Processing} and \textbf{High-Quality Processing}. The raw data encompasses a wide array of Chinese corpora, including news, social media, and blogs, thereby enhancing the coverage and representativeness of our dataset. Following the Fundamental Processing steps, we obtain the CCI3.0 dataset. This dataset undergoes further refinement through model-based High-Quality Processing, resulting in the CCI3.0-HQ dataset. The subsequent sections provide a detailed explanation of these two stages in the dataset construction workflow.

\subsection{Fundamental Processing}
This section outlines the four key processes involved in the Fundamental Processing phase, which is critical for preparing the CCI3.0 dataset and supports subsequent data preparation stages.

\begin{itemize} 
\item \textbf{Safety Filtering}: We implement filters to exclude data from websites likely to contain unsafe content, targeting domains identified as harmful by safety standards and those known for adult material, thereby ensuring compliance with stringent safety criteria. 
\item \textbf{Text Extraction and Cleaning}: Given the dataset's diverse sources, we design specialized parsers for each source to effectively extract and clean the content. 
\item \textbf{Document-Level De-Duplication}: We utilize global MinHash \cite{10.5555/829502.830043} to identify and remove near-duplicate documents, ensuring the dataset's diversity and avoiding redundancy. 
\item \textbf{Heuristic and Basic Quality Filtering}: A set of heuristics is employed to filter out low-quality documents, eliminate outliers, and reduce excessive repetition. Subsequently, we apply a basic quality classifier based on ChineseWebText \cite{chen2023chinesewebtext}, which predicts the likelihood of a text being referenced by reliable sources, such as Wikipedia, e-books, and news articles. \end{itemize}

\subsection{High-quality Processing}
This stage focuses on the distillation of the quality scoring ability of Qwen2-72B-instruct into a 0.5B model to effectively scoring large amount of data. 

\subsubsection{Methods for High-Quality Sample Annotation}
A primary focus of High-Quality Processing is the precise definition of "high quality" within the pre-training context. After exploring and comparing leading methods, we adopt the FineWeb-edu approach to define high-quality samples and develop a classifier targeting high-quality educational content in Chinese. This aims to enhance the overall quality of Chinese corpora. A detailed comparison of annotation methods and their effectiveness is presented in Section \ref{sec:annotation_cmp}.

With quality criteria established, the next challenge is efficiently constructing billions of compliant samples. To address this, we implement a structured process for defining and annotating samples according to established benchmarks, ensuring alignment with necessary educational and informational value—critical for robust Chinese language datasets. The workflow is as follows:

We utilize Qwen2-72B-Instruct to score 145,000 random web samples from the CCI3.0 dataset on a scale from 0 (non-educational) to 5 (highly educational), employing a prompt similar to FineWeb-edu. The locally deployed API of Qwen2, integrated with vLLM \cite{kwon2023efficientmemorymanagementlarge}, facilitates the annotation process. Finally, we perform manual and GPT-4 evaluations on a subset of the labeled results, achieving an agreement rate exceeding 80\%.

\subsubsection{Efficient Training of Quality Classifiers}
Labeling all samples for quality identification with a large model like Qwen2-72B-Instruct would be prohibitively costly. Following the FineWeb-edu methodology, we accumulate hundreds of thousands of annotated samples through automated processes and subsequently train a smaller classification model for efficient labeling at scale. This approach significantly reduces costs while ensuring proper identification of high-quality samples, facilitating comprehensive dataset annotation with practical resource investment.

\begin{figure*}[t]
    \centering
    \subfigure[Effect of Locking vs. Unlocking the Backbone]{
        \includegraphics[width=0.48\linewidth]{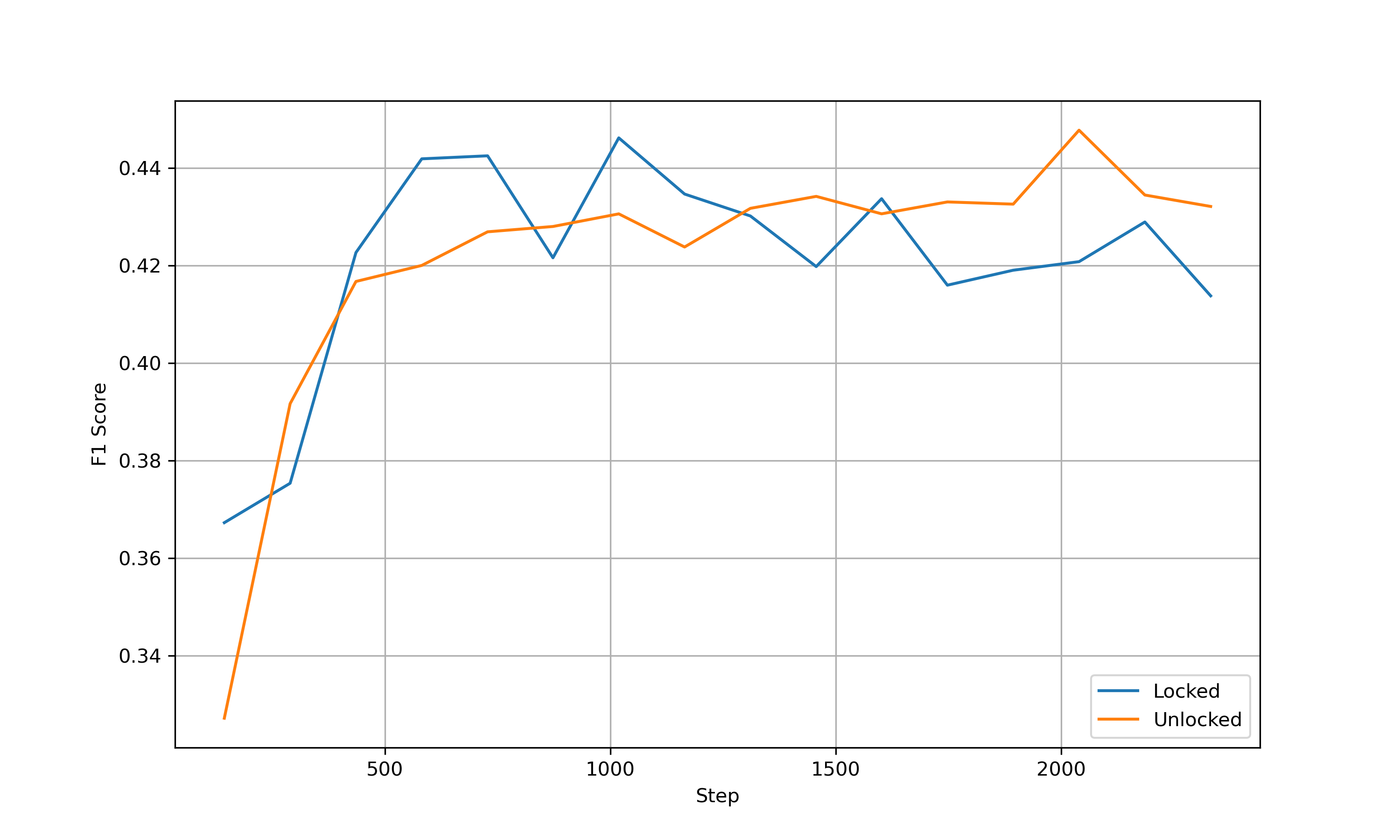}
    }
    \subfigure[Effect of Different Learning Rates]{
        \includegraphics[width=0.48\linewidth]{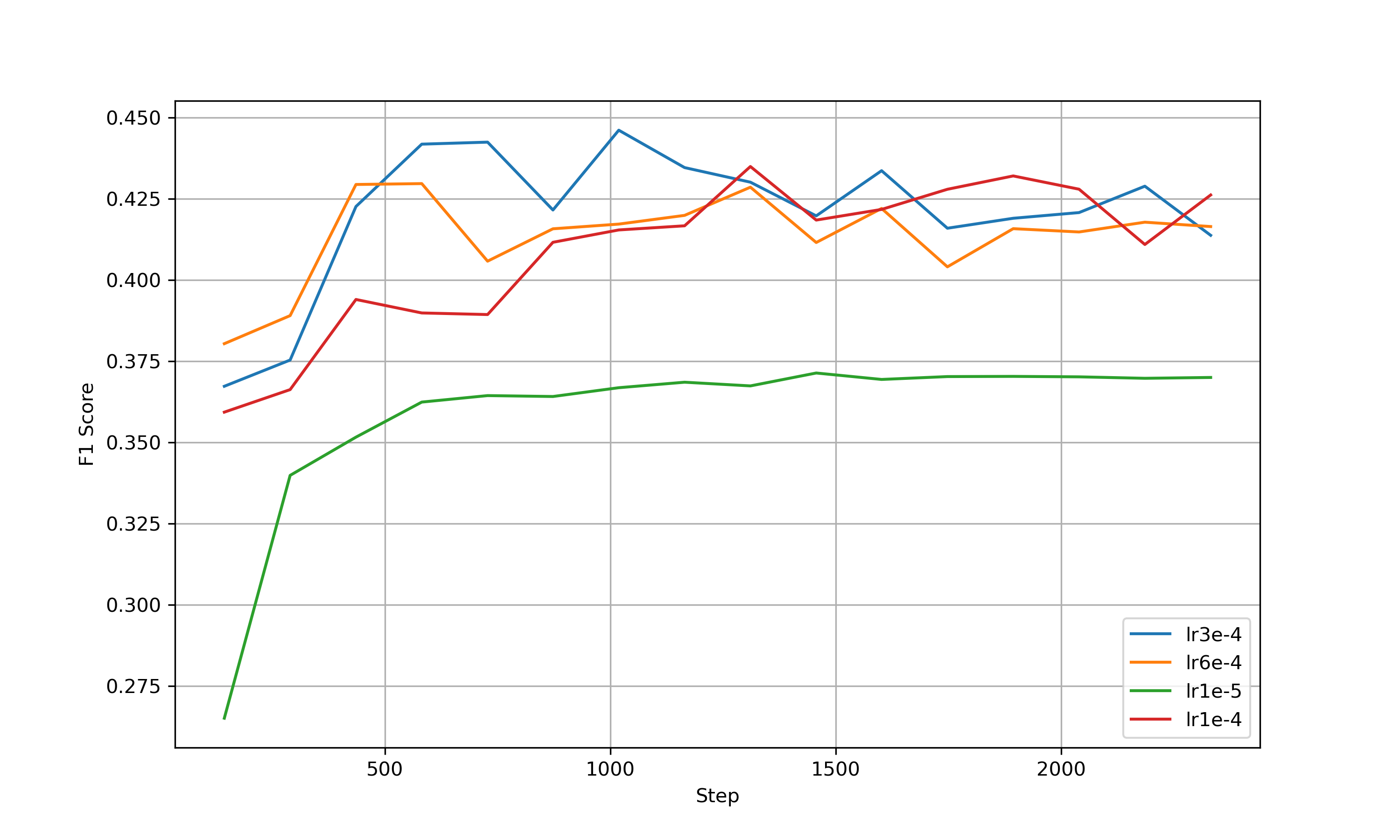}
    }
    \caption{Effects of Backbone Freezing and Learning Rate Adjustments on Classifier Tuning Performance}
    \label{fig:classifier_tuning}
\end{figure*}

We enhance BGE-M3 \cite{chen2024bgem3embeddingmultilingualmultifunctionality} (approximately 0.5B parameters) by adding a classification head with a single regression output, training for 20 epochs at a learning rate of 3e-4. During training, the embedding and encoder layers remain frozen to concentrate on the classification head, with dropout not employed. The training script is available on GitHub\footnote{\url{https://github.com/FlagAI-Open/FlagAI/tree/master/examples/CCI3-HQ}}. The optimal learning rate and intermediate checkpoint are determined based on the F-score across both categories, with training curves documented. For the configuration regarding whether the backbone model is locked in Figure \ref{fig:classifier_tuning}, we chose to lock the backbone model, as the performance improvement when it was not frozen was minimal and choosing to lock the backbone model will save a significant amount of training time. Considering the model's generalization ability, we decided to keep the backbone model in a frozen state. Additionally, we performed a grid search for the learning rate in Figure \ref{fig:classifier_tuning}.

Finally, the model is converted to a binary classifier using a score threshold of 3 and we apply the classifier to about 1.5 billion samples, a process that requires 9700 A100 GPU hours.

\section{Experiments}
In this section, we first conduct experiments to evaluate the effectiveness of our curated corpus in pre-training from scratch. Next, we explore and compare two methods for high-quality annotation, detailing our choice of the FineWeb-edu approach for defining and annotating high-quality samples within the Chinese corpus context. Finally, we perform a comparative analysis of existing high-quality classifiers, highlighting the superior performance of our trained classifier.

\subsection{Experimental Setting}
\subsubsection{Model Training Configuration}
For our evaluation, we utilized the Qwen2-0.5B tokenizer and model architecture, training on a bilingual dataset comprising 100 billion tokens. This configuration ensures effective handling of both Chinese and English data while maintaining experimental consistency. Key training parameters include a sequence length of 4096, a weight decay of 0.1, and gradient clipping at 1.0. The training set comprises 25 million samples with a global batch size of 1024. The learning rate starts at 3e-04, with a minimum of 3e-05 and a warmup covering 2,048,000 samples, following a cosine decay schedule.

\begin{table}[ht]
\centering
\label{tab:model-config}
\begin{tabular}{|l|c|}
\hline
Parameter & Value \\
\hline
attention\_dropout & 0.0 \\
bos\_token\_id & 151849 \\
eos\_token\_id & 151850 \\
hidden\_act & silu \\
hidden\_size & 896 \\
intermediate\_size & 2432 \\
max\_position\_embeddings & 4096 \\
num\_attention\_heads & 14 \\
num\_hidden\_layers & 24 \\
num\_key\_value\_heads & 2 \\
pad\_token\_id & 151643 \\
rms\_norm\_eps & 1e-06 \\
rope\_theta & 10000 \\
tie\_word\_embeddings & True \\
torch\_dtype & bfloat16 \\
vocab\_size & 151851 \\
\hline
\end{tabular}
\vspace{10pt}
\caption{Pre-training Model Configuration Parameters}
\end{table}

\subsubsection{Dataset Composition}
We conducted two primary experiments to evaluate dataset performance:

\begin{itemize}
    \item \textbf{Mixed Dataset Experiment}: This dataset includes 60\% English, 10\% code, and 30\% Chinese content. For the English portion, we employed the FineWeb-edu\footnote{\url{https://huggingface.co/datasets/HuggingFaceFW/fineweb-edu}}, while the code data was sourced from StarCoder\cite{li2023starcodersourceyou}.
    
    \item \textbf{Chinese Dataset Experiment}: This experiment utilized a 100\% Chinese content dataset, incorporating Wanjuan-v1\footnote{\url{https://github.com/opendatalab/WanJuan1.0}}, SkyPile\footnote{\url{https://huggingface.co/datasets/Skywork/SkyPile-150B}}, CCI3.0, and CCI3.0-HQ. The CCI3.0 dataset serves as a baseline, as it has not undergone the high-quality filtering process, allowing for a direct evaluation of the impact of quality improvements on dataset integrity.
\end{itemize}

\subsubsection{Evaluation Metrics}
We employed the lighteval \cite{lighteval} library for model evaluation, mirroring the setup used with the FineWeb dataset and all evaluation metrics are based on a zero-shot setting. Evaluation metrics include:
\begin{itemize}
\item \textit{Average}$_{\textit{Chinese}}$ : Average score of Chinese metrics, including CEval\cite{huang2023ceval} and CMMLU\cite{li2024cmmlumeasuringmassivemultitask}.
\item \textit{Average}$_{\textit{English}}$ : Average score across standard English metrics such as ARC-C\cite{clark2018thinksolvedquestionanswering}, ARC-E\cite{clark2018thinksolvedquestionanswering}, HellaSwag\cite{zellers2019hellaswagmachinereallyfinish}, Winograd\cite{10.5555/3031843.3031909}, MMLU\cite{hendrycks2021measuringmassivemultitasklanguage}, OpenbookQA\cite{banerjee-etal-2019-careful}, PIQA\cite{bisk2019piqareasoningphysicalcommonsense} and SIQA\cite{sap2019socialiqacommonsensereasoningsocial}.
\item \textit{Average}: Combined average score of all evaluation metrics above.
\end{itemize}
Additionally, we compared our classifier against existing models using the F1-score derived from the same test samples, employing the macro-average F1-score from `sklearn.metrics.classification\_report' to quantify performance differences.

\subsection{Impacts of CCI3.0-HQ on Model Training}

We conducted a direct comparison of different datasets through end-to-end pre-training experiments, using the final checkpoint of model training for performance evaluation. Detailed experimental results, including dataset comparisons and metric details, are provided in \ref{dataset_exp}. In analyzing the dataset experiment results, three key points emerge that highlight the performance differences across the various datasets and the strengths of CCI3.0-HQ:

\begin{itemize}
    \item \textbf{Mixed Dataset Experiment Results}: In the mixed dataset evaluations, CCI3.0-HQ consistently performs well across most metrics. Notably, it achieves the highest scores in ARC-E (0.542), Winograd (0.523), MMLU (0.292), and SIQA (0.394), showcasing its robust performance in these specific tasks. CCI3.0, while strong in \textit{HellaSwag} (0.36), is overall outperformed by CCI3.0-HQ. Skypile exhibits good results in \textit{OpenbookQA} (0.334), but lags behind in other metrics, demonstrating a less balanced performance across the board. The ARC-C metric focuses on more challenging knowledge and questions and to address the current gaps in performance for this metric, in future work we plan to increase the quota for high-education-level content to improve results.
    
    \item \textbf{Chinese Dataset Experiment Results}: In the Chinese-specific evaluations, CCI3.0-HQ stands out significantly, particularly in ARC-C (0.235), ARC-E (0.388), and CEval (0.331). These scores surpass all other datasets, solidifying its advantage in tasks involving the Chinese language. Skypile, while performing well in \textit{Winograd} (0.49) and \textit{OpenbookQA} (0.254), is generally less effective in other areas, while Wanjuan-v1 and CCI3.0 trail behind across most metrics.
    
    \item \textbf{Average Performance}: CCI3.0-HQ emerges as the top performer with an overall average score of 0.395, compared to 0.388 for Skypile and CCI3.0. In both English (\textit{Average\textsubscript{English}} = 0.418) and Chinese (\textit{Average\textsubscript{Chinese}} = 0.303) tasks, it maintains a clear lead, affirming its superior generalization across diverse benchmarks. This underlines the effectiveness of the CCI3.0-HQ dataset in enhancing language model training across multilingual tasks.
\end{itemize}

In conclusion, the results clearly demonstrate that CCI3.0-HQ consistently outperforms the other datasets, particularly in key tasks across both English and Chinese benchmarks, making it a superior dataset for comprehensive multilingual evaluation and model training. We also compare the evaluation performance on each dataset at various intermediate checkpoints during training, as detailed in Appendix \ref{sec:training-dynamics}.

\begin{table}[h]
\centering
\label{tab:mix-exp-results}

\begin{tabular}{ccccc}
\hline
\multicolumn{5}{c}{Mixed Dataset Experiment Results} \\
\hline
Metrics & SkyPile & Wanjuan-v1 & CCI3.0 & CCI3.0-HQ \\
\hline
ARC-C & 0.270 & \textbf{0.277} & 0.265 & 0.269 \\
ARC-E & 0.521 & 0.517 & 0.539 & \textbf{0.542} \\
HellaSwag & 0.355 & 0.347 & \textbf{0.36} & 0.357 \\
Winograd & 0.507 & 0.502 & 0.498 & \textbf{0.523} \\
MMLU & 0.286 & 0.287 & 0.289 & \textbf{0.292} \\
OpenbookQA & \textbf{0.334} & 0.312 & 0.326 & 0.318 \\
PIQA & 0.651 & 0.651 & \textbf{0.652} & 0.648 \\
SIQA & 0.38 & 0.387 & 0.375 & \textbf{0.394} \\
\hline
CEval & 0.279 & 0.275 & 0.278 & \textbf{0.296} \\
CMMLU & 0.294 & 0.286 & 0.292 & \textbf{0.309} \\\hline
\textit{Average}$_{\textit{English}}$ & 0.413 & 0.410 & 0.413 & \textbf{0.418} \\
\textit{Average}$_{\textit{Chinese}}$ & 0.287 & 0.280 & 0.285 & \textbf{0.303} \\
\textit{Average} & 0.388 & 0.384 & 0.388 & \textbf{0.395} \\
\hline
\multicolumn{5}{c}{Chinese Dataset Experiment Results} \\
\hline
Metrics & SkyPile & Wanjuan-v1 & CCI3.0 & CCI3.0-HQ \\
\hline
ARC-C & 0.192 & 0.217 & 0.202 & \textbf{0.235} \\
ARC-E & 0.313 & 0.282 & 0.323 & \textbf{0.388} \\
HellaSwag & 0.279 & 0.269 & 0.283 & \textbf{0.295} \\
Winograd & \textbf{0.490} & 0.487 & 0.485 & 0.481 \\
MMLU & 0.244 & 0.254 & 0.245 & \textbf{0.259} \\
OpenbookQA & \textbf{0.254} & 0.232 & 0.232 & 0.242 \\
PIQA & 0.528 & 0.539 & 0.53 & \textbf{0.556} \\
SIQA & \textbf{0.387} & 0.377 & 0.372 & 0.382 \\
\hline
CEval & 0.305 & 0.279 & 0.294 & \textbf{0.331} \\
CMMLU & 0.304 & 0.298 & 0.296 & \textbf{0.328} \\\hline
\textit{Average}$_{\textit{English}}$ & 0.336 & 0.332 & 0.334 & \textbf{0.355} \\
\textit{Average}$_{\textit{Chinese}}$ & 0.304 & 0.289 & 0.295 & \textbf{0.329} \\
\textit{Average} & 0.330 & 0.324 & 0.326 & \textbf{0.350} \\
\hline
\end{tabular}
\vspace{10pt}
\caption{Comparison of Dataset Impacts on Model Performance in Mixed and Chinese Dataset Experiments}
\label{dataset_exp}
\end{table}

\subsection{Assessment of Quality Annotation Techniques}
\label{sec:annotation_cmp}
In the context of pre-training large-scale natural language models, we explore two leading methods for defining high-quality samples: FineWeb-edu and DataComp-LM(DCLM)\cite{li2024datacomplmsearchgenerationtraining}. These two methods are based on a new approach that has recently emerged for filtering pre-training datasets of large language models: using synthetic data to develop classifiers for identifying content. 

\begin{itemize}
    \item \textbf{FineWeb-edu}: FineWeb-edu evaluates the educational quality of web pages on a scale from 0 to 5, focusing primarily on content aimed at grade-school and middle-school levels. To ensure a balance between educational content of various complexities, a threshold of 3 is used during the filtering process. This allows the retention of not only mid-level educational pages but also some high-level content, ensuring that both foundational and advanced knowledge is included in the dataset for further analysis or training purposes.
    \item \textbf{DCLM}: In the DCLM paper, multiple datasets are compared to identify high-quality samples, ultimately selecting the OpenHermes 2.5\cite{OpenHermes2.5} dataset as the positive data. This dataset is then used to train a binary classifier. The trained classifier is subsequently applied to the pre-training corpus to identify high-quality content. Since a Chinese version of the OpenHermes 2.5 dataset is not available, the original dataset is translated into Chinese for use in further processing\footnote{\url{https://huggingface.co/datasets/ldwang/OpenHermes-2.5-zh}}.
\end{itemize}

Through experimental comparisons \ref{tab:fineweb-edu-dclm-cmp}, we find that for Chinese corpora, the FineWeb-edu approach outperformed DCLM especially in \textit{Average}$_{\textit{Chinese}}$. The table highlights the performance comparison between two quality annotation methods: \textit{DCLM} and \textit{FineWeb-edu}. Based on the metrics shown, two key points emerge:

\begin{itemize}
    \item \textbf{Performance in Chinese-Specific Metrics}: In the Chinese-specific metrics, \textit{FineWeb-edu} consistently outperforms \textit{DCLM}. Specifically, for the \textit{CEval} and \textit{CMMLU} benchmarks, \textit{FineWeb-edu} scores 0.331 and 0.328 respectively, whereas \textit{DCLM} trails behind with scores of 0.298 and 0.311. Additionally, the \textit{Average\textsubscript{Chinese}} score shows that \textit{FineWeb-edu} performs significantly better, with a score of 0.329 compared to \textit{DCLM}'s 0.305.
    
    \item \textbf{Overall Performance}: In terms of the overall average, \textit{FineWeb-edu} maintains a slight edge with a score of 0.350 over \textit{DCLM}'s 0.348. While the differences in English-focused metrics are minor, \textit{FineWeb-edu}'s stronger performance in Chinese benchmarks, particularly \textit{CEval} and \textit{CMMLU}, demonstrates its superior adaptability for multilingual evaluation.
\end{itemize}

In summary, the \textit{FineWeb-edu} method shows stronger results in Chinese-specific evaluations, and its overall performance demonstrates its effectiveness, particularly in tasks requiring higher precision in Chinese language datasets. As a result, we decide to adopt the FineWeb-edu method for identifying high-quality samples in the subsequent steps.

\begin{table}[h]
\centering
\begin{tabular}{ccc}
\hline
Metrics & DCLM & FineWeb-edu \\
\hline
ARC-C & 0.211 & \textbf{0.235} \\
ARC-E & 0.378 & \textbf{0.388} \\
HellaSwag & \textbf{0.310} & 0.295 \\
Winograd & \textbf{0.485} & 0.481 \\
MMLU & \textbf{0.259} & 0.259 \\
OpenbookQA & \textbf{0.262} & 0.242 \\
PIQA & \textbf{0.571} & 0.556 \\
SIQA & \textbf{0.389} & 0.382 \\
\hline
CEval & 0.298 & \textbf{0.331} \\
CMMLU & 0.311 & \textbf{0.328} \\ \hline
\textit{Average}$_{\textit{English}}$ & \textbf{0.358} & 0.355 \\
\textit{Average}$_{\textit{Chinese}}$ & 0.305 & \textbf{0.329} \\
\textit{Average} & 0.348 & \textbf{0.350} \\
\hline
\end{tabular}
\vspace{10pt}
\caption{Comparison of Two Quality Annotation Methods}
\label{tab:fineweb-edu-dclm-cmp}
\end{table}

\subsection{Evaluation of Quality Classifiers for Chinese Web data}

All models are converted into binary classifiers using a score threshold of 3.0 and are evaluated on the same test dataset of about 14k samples. These 14k samples were randomly extracted from a large corpus of Chinese texts, containing both the original text and corresponding labels. They can form a benchmark\footnote{\url{https://huggingface.co/datasets/BAAI/CCI3-HQ-Annotation-Benchmark}} for subsequent evaluation of text quality. Our high-quality classifier achieves the best performance and detailed results are in Table \ref{classifier_cmp}. The table provides a comparison of four classifiers: \textit{classifier}$_{\textit{FineWeb-edu}}$, \textit{classifier}$_{\textit{IndustryCorpus2}}$ and \textit{classifier}$_{\textit{ChineseWebText}}$, and our \textit{classifier}$_{\textit{CCI3.0-HQ}}$, evaluated across precision, recall, and F1-score for both positive and negative classes, along with macro averages. Three key observations can be made:

\begin{itemize}
    \item \textbf{Performance on Positive Samples}: The \textit{classifier}$_{\textit{CCI3.0-HQ}}$ classifier demonstrates a notable advantage in classifying positive samples, achieving a precision of 0.86 and an F1-score of 0.53. In contrast, \textit{classifier}$_{\textit{FineWeb-edu}}$, despite its high precision (0.91), has a recall of only 0.02, leading to a very low F1-score of 0.03 for positive samples. This highlights \textit{classifier}$_{\textit{CCI3.0-HQ}}$'s balanced ability to handle positive classes effectively.
    
    \item \textbf{Macro Average Performance}: The macro average F1-score for \textit{classifier}$_{\textit{CCI3.0-HQ}}$ is 0.73, the highest among all classifiers, showing that it maintains strong overall performance across both positive and negative classes. In comparison, \textit{classifier}$_{\textit{FineWeb-edu}}$ has a macro average F1-score of 0.47, largely due to its poor performance on positive samples.
    
    \item \textbf{Balanced Classification}: While \textit{classifier}$_{\textit{IndustryCorpus2}}$ performs well in terms of recall for positive samples (0.86), its lower precision (0.32) results in a lower F1-score for positive classification (0.47). On the other hand, \textit{classifier}$_{\textit{CCI3.0-HQ}}$ balances both precision and recall across classes, achieving a more consistent and reliable performance overall, especially in terms of the macro F1-score (0.73), which suggests that it is the most robust classifier across all datasets.

    \item \textbf{Importance of Quality Classification}: When compared with the \textit{classifier}${\textit{ChineseWebText}}$, the new added \textit{classifier}$_{\textit{CCI3.0-HQ}}$ showcases significantly improved performance in handling diverse data and distinguishing high-quality content. This advancement underscores the essential role that proper quality filtering plays in pre-training, which is also a critical factor contributing to the CCI3.0-HQ dataset’s superior performance over the original CCI3.0 dataset. 
    
\end{itemize}

In summary, \textit{classifier}$_{\textit{CCI3.0-HQ}}$ demonstrates superior classification performance, particularly in handling positive samples more effectively and maintaining a strong macro average across all metrics. Compared to the \textit{classifier}$_{\textit{FineWeb-edu}}$ mainly trained on English corpora, \textit{classifier}$_{\textit{IndustryCorpus2}}$ and \textit{classifier}$_{\textit{ChineseWebText}}$ trained on Chinese corpora, our classifier shows significant improvements in both precision and recall for distinguishing positive and negative samples. We attribute this to better suitability for Chinese and data distribution, as well as the amount of training data and model tuning.

\begin{table}[ht]
    \centering
    \begin{tabular}{l|ccc}
        \hline
        Classifier & Precision & Recall & F1-score \\
        \hline
        \multicolumn{4}{c}{\textit{classifier}$_{\textit{FineWeb-edu}}$} \\
        \hline
        Positive & 0.91 & 0.02 & 0.03 \\
        Negative & 0.82 & 1.00 & 0.90 \\
        Macro F1 & 0.87 & 0.51 & 0.47 \\
        \hline
        \multicolumn{4}{c}{\textit{classifier}$_{\textit{ChineseWebText}}$} \\
        \hline
        Positive & 0.18 & 0.58 & 0.27 \\
        Negative & 0.80 & 0.38 & 0.52 \\
        Macro F1 & 0.49 & 0.48 & 0.39 \\
        \hline
        \multicolumn{4}{c}{\textit{classifier}$_{\textit{IndustryCorpus2}}$} \\
        \hline
        Positive & 0.32 & 0.86 & 0.47 \\
        Negative & 0.95 & 0.59 & 0.73 \\
        Macro F1 & 0.64 & 0.73 & 0.60 \\
        \hline
        \multicolumn{4}{c}{\textit{classifier}$_{\textit{CCI3.0-HQ}}$} \\
        \hline
        Positive & 0.86 & 0.38 & \textbf{0.53} \\
        Negative & 0.88 & 0.99 & \textbf{0.93} \\
        Macro F1 & 0.87 & 0.68 & \textbf{0.73} \\
        \hline   
    \end{tabular}
    \vspace{10pt}
    \caption{Evaluation of Different Quality Classifiers}
    \label{classifier_cmp}
\end{table}

\section{Conclusion and Limitation}

We have released and open-sourced the CCI3.0-HQ dataset, which has undergone sophisticated hybrid quality filtering methodology to enhance data integrity. Through comparison after pre-training small-scale models from scratch and rigorous experimental evaluations, CCI3.0-HQ significantly outperforms existing well-known Chinese open-source datasets. Also, We introduce and open-source the CCI3-HQ classifier, which demonstrates superior performance compared to existing open-source Chinese and English quality classifiers and the CCI3.0-HQ dataset demonstrates the importance of high-quality filtering in the pre-training of Chinese large language models. As the largest high-quality Chinese pre-training corpus currently available, CCI3.0-HQ is poised to contribute to the advancement of large language models, especially those focused on Chinese. 
The dataset includes data collected up until early 2024, which means it might lack information about more recent events or trends. Despite the data cleaning efforts to enhance the dataset’s quality, there may still be some lower-quality samples present. We will continue data processing and quality filtering efforts to further support the development of high-quality large language models. As future work, the Infinity Instruct\footnote{\url{https://huggingface.co/datasets/BAAI/Infinity-Instruct}} dataset can be leveraged to further optimize the quality classifier, which will lead to additional improvements in the performance of the Aquila series of large language models\cite{zhang2024aquila2technicalreport}\cite{zhang2024aquilamoeefficienttrainingmoe}.







\bibliographystyle{unsrt}  
\bibliography{references}

\section{Appendix}

\begin{figure}[htbp]
    \centering
    \begin{minipage}[t]{0.49\linewidth}
        \centering
        \begin{tikzpicture}
            \begin{axis}[
                width=\textwidth,
                height=0.6\textwidth,
                xlabel={Training Tokens (B)},
                ylabel=\textit{Average},
                legend pos=south east,
                grid=major,
                xmin=0, xmax=100,
                ymin=0.35, ymax=0.4,
                xtick={0,20,40,60,80,100},
                ytick={0.36,0.37,0.38,0.39,0.4},
                legend style={font=\tiny},
                legend cell align={left}
            ]
            \addplot+[mark=*, color=blue] coordinates {
                (0, 0.3623)
                (20, 0.3676)
                (40, 0.3776)
                (60, 0.3808)
                (80, 0.3860)
                (100, 0.3841)
            };
            \addlegendentry{Wanjuan-v1}

            \addplot+[mark=triangle*, color=red] coordinates {
                (0, 0.3623)
                (20, 0.3695)
                (40, 0.3796)
                (60, 0.3834)
                (80, 0.3865)
                (100, 0.3875)
            };
            \addlegendentry{CCI3.0}

            \addplot+[mark=square*, color=green] coordinates {
                (0, 0.3623)
                (20, 0.3717)
                (40, 0.3874)
                (60, 0.3906)
                (80, 0.3947)
                (100, 0.3948)
            };
            \addlegendentry{CCI3.0-HQ}

            \addplot+[mark=o, color=purple] coordinates {
                (0, 0.3623)
                (20, 0.3666)
                (40, 0.3768)
                (60, 0.3764)
                (80, 0.3805)
                (100, 0.3877)
            };
            \addlegendentry{SkyPile}

            \end{axis}
        \end{tikzpicture}
        \caption{Mixed Dataset Experiment}
        \label{fig:mix-exp-cmp}
    \end{minipage}
    \hfill
    \begin{minipage}[t]{0.49\linewidth}
        \centering
        \begin{tikzpicture}
            \begin{axis}[
                width=\textwidth,
                height=0.6\textwidth,
                xlabel={Training Tokens (B)},
                ylabel=\textit{Average}$_{\textit{Chinese}}$,
                legend pos=south east,
                grid=major,
                xmin=0, xmax=100,
                ymin=0.25, ymax=0.36,
                xtick={0,20,40,60,80,100},
                ytick={0.26,0.28,0.30,0.32,0.34,0.36},
                legend style={font=\tiny},
                legend cell align={left}
            ]

            \addplot+[mark=*, color=blue] coordinates {
                (0, 0.2550)
                (20, 0.3179)
                (40, 0.3237)
                (60, 0.3244)
                (80, 0.3226)
                (100, 0.3235)
            };
            \addlegendentry{Wanjuan-v1}

            \addplot+[mark=triangle*, color=red] coordinates {
                (0, 0.2550)
                (20, 0.3187)
                (40, 0.3214)
                (60, 0.3243)
                (80, 0.3270)
                (100, 0.3262)
            };
            \addlegendentry{CCI3.0}

            \addplot+[mark=square*, color=green] coordinates {
                (0, 0.2550)
                (20, 0.3361)
                (40, 0.3403)
                (60, 0.3476)
                (80, 0.3528)
                (100, 0.3497)
            };
            \addlegendentry{CCI3.0-HQ}

            \addplot+[mark=o, color=purple] coordinates {
                (0, 0.2550)
                (20, 0.3234)
                (40, 0.3233)
                (60, 0.3258)
                (80, 0.3265)
                (100, 0.3296)
            };
            \addlegendentry{SkyPile}

            \end{axis}
        \end{tikzpicture}
        \caption{Chinese Dataset Experiment}
        \label{fig:zh-exp-cmp}
    \end{minipage}
\end{figure}

\subsection{Evaluation of Training Dynamics}
\label{sec:training-dynamics}
We conduct evaluations at every 20 billion tokens of training to closely monitor and compare the performance of various datasets throughout the training process. All results across training tokens of \textbf{Mixed Dataset Experiment} and \textbf{Chinese Dataset Experiment} are depicted in Figures \ref{fig:mix-exp-cmp} and \ref{fig:zh-exp-cmp}. The figures compare the performance of different datasets across training tokens in both mixed and Chinese-specific dataset experiments. Three key points emerge from the analysis:

\begin{itemize}
    \item \textbf{Superior Performance of CCI3.0-HQ}: In both the mixed dataset experiment and the Chinese dataset experiment, \textit{CCI3.0-HQ} consistently demonstrates superior performance compared to the other datasets. In the mixed experiment, \textit{CCI3.0-HQ} achieves the highest average score of 0.395 when trained with 100B tokens, significantly outperforming \textit{Wanjuan-v1}, \textit{CCI3.0}, and \textit{SkyPile}. Similarly, in the Chinese-specific experiment, \textit{CCI3.0-HQ} leads with a score of 0.355, showcasing its robustness in both multilingual and Chinese-centric tasks.
    
    \item \textbf{Gradual Improvement Across Tokens}: All datasets show an increase in performance as the number of training tokens increases. However, \textit{CCI3.0-HQ} demonstrates a steeper improvement curve in both experiments, indicating that it scales more efficiently with larger amounts of training data. In contrast, \textit{Wanjuan-v1} and \textit{SkyPile} show slower growth, particularly in the Chinese dataset experiment.
    
    \item \textbf{Mixed Dataset vs. Chinese Dataset Performance}: While all datasets perform better in the mixed dataset experiment compared to the Chinese-specific one, \textit{CCI3.0-HQ} maintains a notable gap over the others in both scenarios. This suggests that \textit{CCI3.0-HQ} has been effectively optimized for both general and Chinese-specific data, making it the most balanced and high-performing dataset overall.

\end{itemize}

The same conclusion confirms that the results highlight the significant advantage of the \textit{CCI3.0-HQ} dataset in both general and Chinese-specific tasks, demonstrating superior scalability and adaptability throughout the training process. The intermediate checkpoints of the models trained in all comparison experiments will be open-sourced \footnote{\url{https://huggingface.co/BAAI/CCI3-HQ-Intermediate-Checkpoints}}.

\end{document}